\documentclass[runningheads]{llncs}
\usepackage{graphicx}
\usepackage{subfigure}
\usepackage{booktabs}
\usepackage{amsmath}
\usepackage{array}

\begin{document}

\title{Contextual Prediction Difference Analysis for Explaining Individual Image Classifications}
%
%
\author{Jindong Gu\inst{1,2} \and Volker Tresp\inst{1,2}}

\authorrunning{Jindong Gu et al.}
\institute{University of Munich, Germany \and
Siemens AG, Corporate Technology, Germany \\
\email{\{jindong.gu, volker.tresp\}@siemens.com}}

\maketitle              
\begin{abstract}
Much effort has been devoted to understanding the decisions of deep neural networks in recent years. A number of model-aware saliency methods were proposed to explain individual classification decisions by creating saliency maps. However, they are not applicable when the parameters and the gradients of the underlying models are unavailable. Recently, model-agnostic methods have also received attention. As one of them, \textit{Prediction Difference Analysis} (PDA), a probabilistic sound methodology, was proposed. In this work, we first show that PDA can suffer from saturated classifiers. The saturation phenomenon of classifiers exists widely in current neural network-based classifiers. To explain the decisions of saturated classifiers better, we further propose Contextual PDA, which runs hundreds of times faster than PDA. The experiments show the superiority of our method by explaining image classifications of the state-of-the-art deep convolutional neural networks.
\end{abstract}

\section{Introduction}
Machine learning enables many recent advances in artificial intelligence. Especially, Deep Neural Networks (DNNs) achieve start-of-the-art performance on many tasks \cite{lecun2015deep}. One of the weaknesses of DNNs is the lack of interpretability. It is difficult to explain the model's decisions to end-users. Furthermore, even machine learning experts also have difficulty in understanding individual decisions of neural networks with deep architectures. They are often applied as black boxes to tackle problems of different domains. Big technology companies like Google, Microsoft, and IBM also offer machine learning-based cloud services, e.g., general vision recognition systems, on which many practitioners have built their applications for end-users.

In real-world applications, however, individual decisions need to be explained to gain trust from the users. E.g., autonomous driving systems should reassure passengers by giving explanations when braking the car abruptly \cite{kim2017interpretable,kim2018textual}. Decisions made by deep models are also required to be verified in the medical domain. Mistakes of unverified models could have an unexpected impact on humans or lead to unfair decisions \cite{liu2018delayed,hashimoto2018fairness}. Besides, AI applications must comply with related legislation, e.g., the right to explanation in GDPR of the European Union \cite{selbst2017meaningful}.

\begin{table}
\begin{center}
\caption{Summarization of different approaches for explaining image classifications.} \label{tab:summary}
\begin{tabular}{ | c | m{8.2cm} | }
\hline
\textbf{Types of Explanation} &  \multicolumn{1}{|c|}{ \textbf{Approaches}} \\
\hline
Saliency Maps & Identifying the relevance of each input pixel to an output class with model-aware saliency methods \cite{Simonyan2013DeepIC,springenberg2014striving,bach2015pixel,selvaraju2017grad,shrikumar2017learning,sundararajan2017axiomatic,smilkov2017smoothgrad,Gu2018UnderstandingID,srinivas2019full,gu2019saliency,gu2019semantics} or model-agnostic ones  \cite{dabkowski2017real,schwab2019cxplain,zeiler2010deconvolutional,ribeiro2016should,zintgraf2017visualizing,fong2017interpretable,lundberg2017unified}. \\
\hline
Explanatory Sentences & Generating natural language sentences that describe the class-discriminative pixels \cite{hendricks2016generating,hendricks2018grounding}.  \\
\hline
Counterfactual Images & Identifies how the given input could change such that the classifier would output a different specified class \cite{Chang2018ExplainingIC,Goyal2019CounterfactualVE}. \\
\hline
Supporting Training Images & Identifying training points most responsible for a given prediction \cite{koh2017understanding}. \\
\hline
\end{tabular}
\end{center}
\end{table}

DNN-based image classifications pose more challenges to identify the relationship between input pixels and output classes since the input space is often high-dimension. In recent years, many directions have been explored to explain individual image classifications, which can be summarized in Table \ref{tab:summary}.

Saliency Maps, as intuitive explanations, have received much attention. Model-aware saliency methods leverage the parameters and the gradients of neural networks to compute saliency maps, while model-agnostic ones only use inputs and outputs. Model-agnostic saliency methods are preferred for at least three reasons: they are able to explain any classifiers; the explanations produced from two or more different types of models are comparable; an ensemble model can be explained without requiring knowledge of model components. 

Model-agnostic saliency methods can be further categorized into two subcategories. Given a predictive model $f_{pred}(\cdot): R^n \mapsto R^k$ and a test instance $\pmb{x}$, the prediction $f_{pred}(\pmb{x})$ needs to be explained. 

1) The approaches of the first subcategory build an explanation model $f_{exp}: R^n \mapsto R^n$ \cite{dabkowski2017real,schwab2019cxplain}. The explanation model is often a neural network with U-net architecture \cite{ronneberger2015u}, which is optimized so that the created explanations $f_{exp}(\pmb{x})$ as accurate as possible on the whole training dataset. In this subcategory, the training data are required to train the explanation model. In addition, it can output a wrong explanation for the test instance $\pmb{x}$, since the explanation model is only optimized to minimize the explanation error on the whole dataset. The mistake is unacceptable in some domains, e.g., in the medical domain. 

2) The approaches of the second subcategory first create different variants of the test instance $\pmb{x}$, make predictions on them using $f_{pred}(\cdot)$, and create explanations using the inputs and the outputs \cite{zeiler2010deconvolutional,ribeiro2016should,zintgraf2017visualizing}. It ensures that the created explanations are faithful to the prediction. This work focuses on the second subcategory of model-agnostic saliency methods. To be noted that LIME can also be categorized into this subcategory. LIME \cite{ribeiro2016should} first creates variants of the input image by perturbating its superpixels, makes predictions on these variants, trains an interpretable classifier, and identifies the relevance of each image superpixels.

Occlusion methods replace a subset of the features with pre-defined values and identify the resulting difference of the target output as the relevance of the features \cite{zeiler2010deconvolutional}. However, the occlusion with pre-defined values cannot remove the evidence of the occluded features without introducing unexpected features \cite{szegedy2013intriguing}. A probabilistic sound methodology, \textit{Prediction Difference Analysis} (PDA), was proposed to overcome this shortcoming by marginalizing the effect of the features \cite{zintgraf2017visualizing}. In this work, we show that PDA suffers from the saturation of classifiers and further propose Contextual PDA.

The next section introduces PDA. Sec. \ref{sec:SaC} shows the saturation of existing classifiers and the ineffectiveness of PDA for saturated classification. In Sec. \ref{sec:CPDA}, we present our contextual PDA and demonstrate its effectiveness. In the experiment section, we evaluate our method. Sec. \ref{sec:related} discusses the related work. The last section provides conclusions and discusses future work.

\section{Prediction Difference Analysis}
\label{sec:PDA}
A classifier is a function mapping input instances into probabilities of output classes $f:\pmb{x} \mapsto f(\pmb{x})$. Each instance has $n$ features $ \pmb{x} = \{x_1, x_2,\cdots, x_n\}$. Given a classification decision $f(\pmb{x})$ of the instance $\pmb{x}$, saliency methods assign a relevance value $r_i$ to each feature $x_i$. The resulted saliency map is defined as $R = \{r_1, r_2,\cdots, r_n\}$. To identify the effect a feature $x_i$ has on the prediction of the instance $\pmb{x}$, PDA computes the difference between two predictions:
\begin{equation}
r_i = D(f(\pmb{x}), f(\pmb{x}_{\backslash i}))
\end{equation}
where $f(\pmb{x}_{\backslash i})$ means the model's prediction without the knowledge of the feature $x_i$ (marginal prediction) and $D()$ is a distance function measuring the difference between two predictions. Two different ways are often used to evaluate such difference: the first is based on the information theory. For a given class $y$, this measure computes the difference of the amount of information necessary to find out that y is true for the given instance $\pmb{x}$.
\begin{equation}
r_i = \log_2 p(y|\pmb{x}) -  \log_2 p(y|\pmb{x}_{\backslash i}) [bit]
\end{equation}
The second one is easy to understand. It directly computes the difference between two probabilities as follows:
\begin{equation}
r_i = p(y|\pmb{x}) -  p(y|\pmb{x}_{\backslash i})
\label{equ:pda}
\end{equation}
In this paper, we evaluate the difference using the second measure. Thereinto the first term $p(y|\pmb{x})$ can be obtained by just classifying the instance $\pmb{x}$ with the classifier to be explained. The second term $p(y|\pmb{x}_{\backslash i})$ is more tricky to compute. The easiest way is to set the feature $x_i$ unknown, which is possible only in few classifiers, e.g., naive Bayesian classifier. Another way to remove the effect of the feature $x_i$ is to retrain the classifier with the feature left out. Such retraining procedure is not feasible when the input instance is high-dimensional, e.g., images. A practical approach to simulate the absence of a feature is to marginalize the feature:
\begin{equation}
p(y|\pmb{x}_{\backslash i}) = \sum^M_{k=1}p(x_i = v_k | \pmb{x}_{\backslash i}) p(y|\pmb{x}_{\backslash i}, x_i = v_k)
\end{equation}
where $v_k$ is a possible value for the feature variable $x_i$. The work \cite{robnik2008explaining} approximates the conditional probability $p(x_j | \pmb{x}_{\backslash i})$ with the prior probability $p(x_j)$ and considers a single input variable as one feature. \cite{zintgraf2017visualizing} focuses on explaining the classifications of images on deep convolutional neural networks, which takes the conditional probability $p(v_k | \pmb{x}_{\backslash i})$ into consideration. They consider rectangular patches of images as individual features, which requires less forward inferences. To further make the analysis more efficient, \cite{tian2017visualizing,wei2018explain} considers super-pixels as individual features.

\section{PDA on Saturated Image Classifications}
\label{sec:SaC}
Assuming that an output class of a model is supported by multiple input features, the output value does not change after removing part of the features. This phenomenon is defined as model saturation in this work. In this section, we first demonstrate that PDA can suffer from the classifier saturation. Given an instance $\pmb{x} = \{x_1, x_2, x_3\}$ and a classifier specifying its probability of belonging to a class c: $p(c|\pmb{x}) = \max(x_1, x_2, x_3)$ where $ x_i \in \{0, 1\}$. We aim to explain individual classification decisions made by the classifier using PDA. A concrete instance $\pmb{x} = \{1, 1, 0\}$ is classified as $p(c|\pmb{x}) = \max(1, 1, 0)=1$. Following the equation \ref{equ:pda} in PDA, the relevance values of the three input features are as follows:
\begin{equation}
r_i = p(c|\pmb{x}) - p(c|\pmb{x}_{\backslash i}) = 0, i  \in \{1, 2, 3\}
\end{equation}
The analysis result means that the relevance values of all the features are zeros, which is contrary to the fact that the features $\{x_1, x_2\}$ support the classification decision.

\begin{figure}[t]
\centerline{\includegraphics[width=0.9\columnwidth]{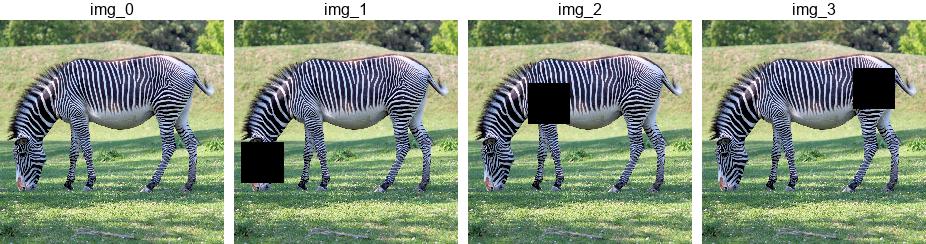}}
\caption{The figure shows a zebra image and its three variants where different visual features are removed.} \label{fig:clip_img}
\end{figure}

The saturation phenomenon exists widely in the current state-of-the-art classifiers. In image classification, many visual features could contribute to a single classification decision at the same time. For instance, in \textit{img\_0} of Figure \ref{fig:clip_img}, the \textit{strip} texture of both two \textit{zebras} support the hypothesis that the image belongs to the class \textit{Zebra}. We analyze the classification decision on the state-of-the-art classifiers using prediction difference. We illustrate the analysis using three samples, in which different parts of the image are removed (see the last three images \textit{img\_1, img\_2, img\_3} in Figure \ref{fig:clip_img}).

Table \ref{tab:dif} shows prediction differences of the respective samples, which is the relevance value of the removed visual features. All the classifiers are confident to classify the original image as a zebra. In image \textit{img\_1}, we remove the head of the zebra and classify the rest of the visual features. The confidence of all classifiers hardly decreases, which means these classification decisions are saturated. Similarly, we remove a part of the body in image \textit{img\_2} and \textit{img\_3}; all the classifier also show high confidence as in the original image. PDA assigns the difference as the relevance of the removed visual features to the classification decision. We can observe that PDA does not work in this case. We also replace the clipped patches with random noise and Gaussian noise, which also show the same conclusion.

\begin{table}[t]
\begin{center}
{\caption{The samples in Figure \ref{fig:clip_img} are classified by the state-of-the-art deep models (Alexnet \cite{krizhevsky2012imagenet}, VGG16 \cite{Simonyan2014VeryDC}, Inception\_V3 \cite{szegedy2016rethinking}, ResNet \cite{he2016deep}, DenseNet \cite{huang2017densely}). The classification scores of zebra class are listed. Their difference to the score of the original image are also shown.} \label{tab:dif}}
\begin{tabular}{lcccc}
\toprule
Image            &  img\_0 & img\_1  & img\_2 & img\_3  \\
\midrule
Alexnet          & .9999  & .9999{\scriptsize (+.0000)}  & .9999{\scriptsize (+.0000)}  & .9999{\scriptsize (+.0000)}  \\
\\[-0.7em]
VGG16           & .9999  & .9995{\scriptsize (-.0004)}   & .9999{\scriptsize (+.0000)}  & .9997{\scriptsize (-.0002)}  \\
\\[-0.7em]
Incept.\_V3 & .9999  & .9999{\scriptsize (+.0000)}      &.9999{\scriptsize (+.0000)}    & .9999{\scriptsize (+.0000)}  \\
\\[-0.7em]
ResNet          & .9985   & .9987{\scriptsize (+.0002)}   & .9975{\scriptsize (-.0010)}  & .9987{\scriptsize (+.0002)} \\
\\[-0.7em]
DenseNet      & .9805   & .9711{\scriptsize (-.0094)}   & .9900{\scriptsize (+.0095)}   & .9873{\scriptsize (+.0068)} \\
\bottomrule
\end{tabular}
\end{center}
\end{table}

One might attribute this saturation phenomenon to the Softmax layer of CNNs. We also conduct further experiments, in which we remove the Softmax layer and make prediction difference analysis using the logits of the ground-truth class. The experimental result shows that the removal of Softmax layer cannot always bypass the saturation phenomenon of CNNs (e.g., in Inception\_V3, ResNet). Furthermore, note that we aim to propose a new \textbf{model-agnostic} explanation method in this work. In the context of the model-agnostic setting, it is impossible to access models. The logits before the softmax layer cannot be accessed if existing. Besides, we even do not know whether a Softmax layer is used or not in the underlying model since the model type (CNNs, SVM \cite{hearst1998support} or other classifiers) are also unknown. Hence, we show the saturation phenomenon of CNNs with softmax layers in this paper.

A similar model saturation phenomenon is also discussed in DeepLIFT \cite{shrikumar2017learning}, where it is also shown that both perturbation-based approaches and gradient-based approaches fail to model saturation. They presented a method with significant advantages over gradient-based methods, in terms of the model saturation. Similar to integrated gradient \cite{sundararajan2017axiomatic}, their method considers the output behaves over a range of inputs instead of the local behavior of the output at the specific input value. The method requires access to model parameters and the activation values of forwarding inferences.

Another limitation of PDA is that the clipped sample could introduce some extra unexpected evidence. In the worst case, adversarial artifacts could be introduced. As observed in many publications \cite{szegedy2013intriguing,goodfellow2001explaining}, the deep neural network is vulnerable to artificial perturbations. Even though the introduction of adversary artifacts rarely happens via sampling in practice, it is an unignorable limitation of the PDA.

\section{Contextual Prediction Difference Analysis}
\label{sec:CPDA}
In the last section, we introduced that PDA could suffer from classifier saturation and possible adversary artifacts. In this section, we propose Contextual PDA (CPDA) to overcome the shortcomings. When a feature $x_i$ is removed from the original instance $\pmb{x}$, the low prediction difference means the feature $x_i$ is not relevant according to the equation \ref{equ:pda} of PDA, namely, $r_i = f(\pmb{x}) - f(\pmb{x}_{\backslash i}) \approx 0 $. 

In our contextual PDA, instead of handling the relevance $r_i$ of a single feature $x_i$ directly, we compute the relevance of contextual information (i.e., the rest of features).
\begin{equation}
\begin{split}
R_{\backslash i} &= f(\pmb{x}) - \sum^M_{k=1}  p(\pmb{x}_{\backslash i} = \pmb{v}_{k} | x_i) p(y|x_i, \pmb{x}_{\backslash i} = {\pmb{v}_{k}}) \\
                            &= f(\pmb{x}) - p(y|x_i) = f(\pmb{x}) - f(x_i)  \\
\end{split}
\label{equ:cpda}
\end{equation}
where $R_{\backslash i}$ are the relevance values of all the features except $x_i$, and $\pmb{v}_{k}$ are possible values of variables $\pmb{x}_{\backslash i}$. $R_{\backslash i}$ can be distributed to individual features in many different ways. For the simplification, we take the equal distribution. 
\begin{equation}
r_i =  \sum_{j \neq i} \frac{R_{\backslash j}}{|R_{\backslash j}|}
\label{equ:equaldis}
\end{equation}
An alternative formulation to identify the relevance of individual attribute is $r_i = f(x_i)$, which computes the effect of a single feature. However, this formulation can not identify negative evidence since $r_i = f(x_i) \geq 0$. The rank of identified relevance values is equivalent to the one when marginalizing all the rest of the features.

\textbf{How can the contextual formulation overcome classifier saturation?} We use the demo introduced in Section \ref{sec:SaC}, in which an instance is classified as a class $c$ with the probability $p(c|\pmb{x}) = \max(1, 1, 0)=1$. With the equation \ref{equ:cpda} and \ref{equ:equaldis}, the analysis of contextual PDA is as follows
\begin{equation}
\begin{split}
r_1 &= \frac{R_{\backslash 2}}{|R_{\backslash 2}|} + \frac{R_{\backslash 3}}{|R_{\backslash 3}|} = 0/2 + 1/2 = 0.5 \\
r_2 &= \frac{R_{\backslash 1}}{|R_{\backslash 1}|} + \frac{R_{\backslash 3}}{|R_{\backslash 3}|} = 0/2 + 1/2 = 0.5 \\
r_3 &= \frac{R_{\backslash 1}}{|R_{\backslash 1}|} + \frac{R_{\backslash 2}}{|R_{\backslash 2}|} = 0/2 + 0/2 = 0 \\
\end{split}
\end{equation}
The features $x_1$ and $x_2$ make equal positive contributions to the decision. The last feature $x_3$ contributes nothing to the decision in this demo. The analysis result illustrates that our contextual PDA can overcome the saturation problem. The reason behind this is that the relevance of each feature depends on inferences of all contextual features instead of the feature itself.

\textbf{How can CPDA be applied to high-dimensional data?} In the following, we introduce the challenges when explaining individual classifications of real-world high-resolution images. It is inefficient to take pixels as individual features since the images are high-dimensional. Additionally, the individual pixels are not human-understandable visual features. Hence, we take image patches as individual features. The shape of image patches we select is squared (not superpixels), which makes the computation of $f(x_i)$ possible. 

The neural network-based classifiers often require a fixed input size, and the selected image patches should be resized to the required size. As the results in Section 6 suggest, our CPDA works fine in practice, but it should be mentioned that the proposed method works the best with classifiers robust to scale and translation. Our method is unsuitable for explaining the neural networks trained on images with a corrected target object, e.g., MNIST digit images.

In addition, not all pixels inside a patch are equally important. The resolution of the obtained saliency maps depends on the size of a single image patch. While too big patches lead to low-resolution saliency maps, too small patches are not visually understandable ($f(x_i)$ makes nonsense). To avoid the dilemma, instead of splitting the image into separated patches, we slide the patch over the whole image with a fixed stride like in convolutional operations. The sensitivity of the stride and the patch size will be further discussed in the experimental section.

\section{Experiment}
\label{sec:exp}
In experiments, we verify the superiority of the proposed CPDA and apply it to explain image classification decisions made by state-of-the-art deep convolutional neural network. In supplementary materials, we also demonstrate the application of our method to commercial general vision recognition systems and understanding the difference between different CNNs.

\subsection{Image Classification with Deep CNNs}
In this experiment, we explain individual image classifications of deep convolutional neural networks. We take pre-trained VGG16 models from pytorch framework as classifiers $f()$, which requires a fixed input size $224 \times 224$. The input image $\pmb{x}$ of size $H \times W$ is resized to the fixed input size, processed and classified as $f(\pmb{x})$. The squared image patch $k \times k$ is clipped from the processed image of $224 \times 224$ and resized to the fixed input size again. The single image patch $x_i$ is classified as $f(x_i)$. We assume that the image classifier is invariant to image translation and scaling. When resizing an image patch, instead of resizing the clipped patch itself, we resize the corresponding patch of the original image to avoid unexpected scaling effects (e.g., too blurry).

\begin{figure}[t]
\centerline{\includegraphics[scale=0.152]{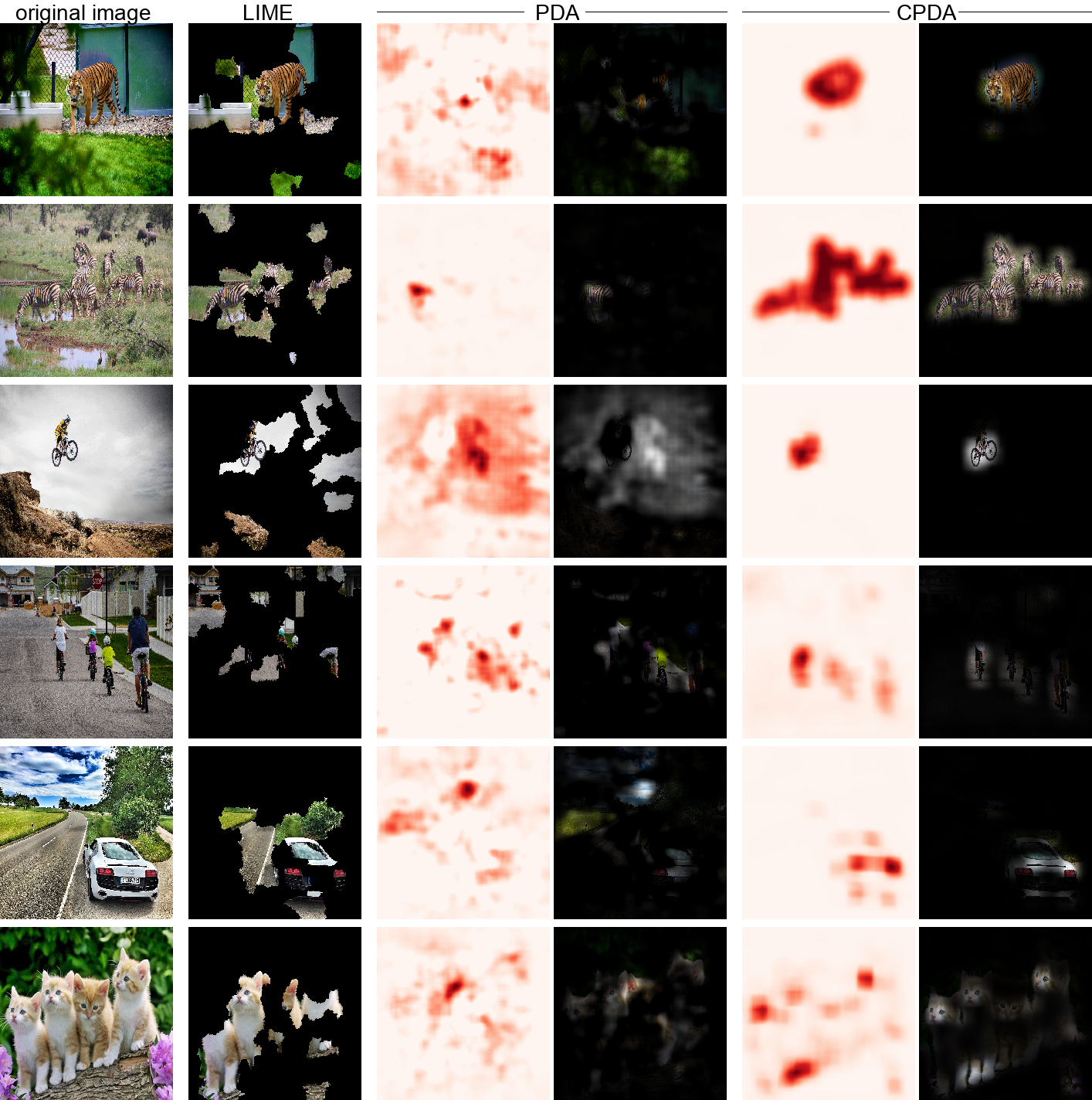}}
\caption{Explanations created by model-agnostic methods.} \label{fig:exp_img}
\end{figure}

\subsubsection{Qualitative and Quantitative Evaluation of Saliency Maps:} We first classify the images using VGG16 and explain each classification using three methods, i.e., LIME, PDA, and our approach CPDA. The explanations of individual classifications are shown in Figure \ref{fig:exp_img}. Each row corresponds to an image. The first column is the original image, and the second column shows the explanation produced by LIME, where the black regions are not relevant to the classification decision, and visible superpixels are relevant visual features. We can find that the explanations of LIME often show many irrelevant features, although it can identify the relevant parts.

We visualize the relevance values of all pixels identified by PDA and CPDA, respectively. In the visualized saliency maps, the red color marks relevant visual features, while the white-color regions are not relevant to the classification decision. To compare them with LIME, we apply the saliency maps as masks on the corresponding original images. The relevance values of saliency maps correspond to the transparency of the mask, which means the irrelevant parts will be blocked. As shown in Figure \ref{fig:exp_img}, our approach CPDA can identify the relevant features more accurately than PDA.

For instance, in the first row of Figure \ref{fig:exp_img}, PDA identifies the grass in the background also as relevant features to the tiger class. Such an error could be caused by the saturation of the classifier. The partial removal of the tiger does not lead to a significant drop in the classification output of the target class. However, when a part of the background is removed, the small drop of classification results indicates the removed part is also relevant. Our approach CPDA finds that the pixels of the tiger are relevant. CPDA identifies the relevance values of visual features according to all the classification results of the contextual information. Hence, the features identified by CPDA are more accurate. Besides, the explanations produced by CPDA is more visually pleasuring than the ones by LIME and PDA.

\begin{figure}[t]
\centerline{\includegraphics[scale=0.15]{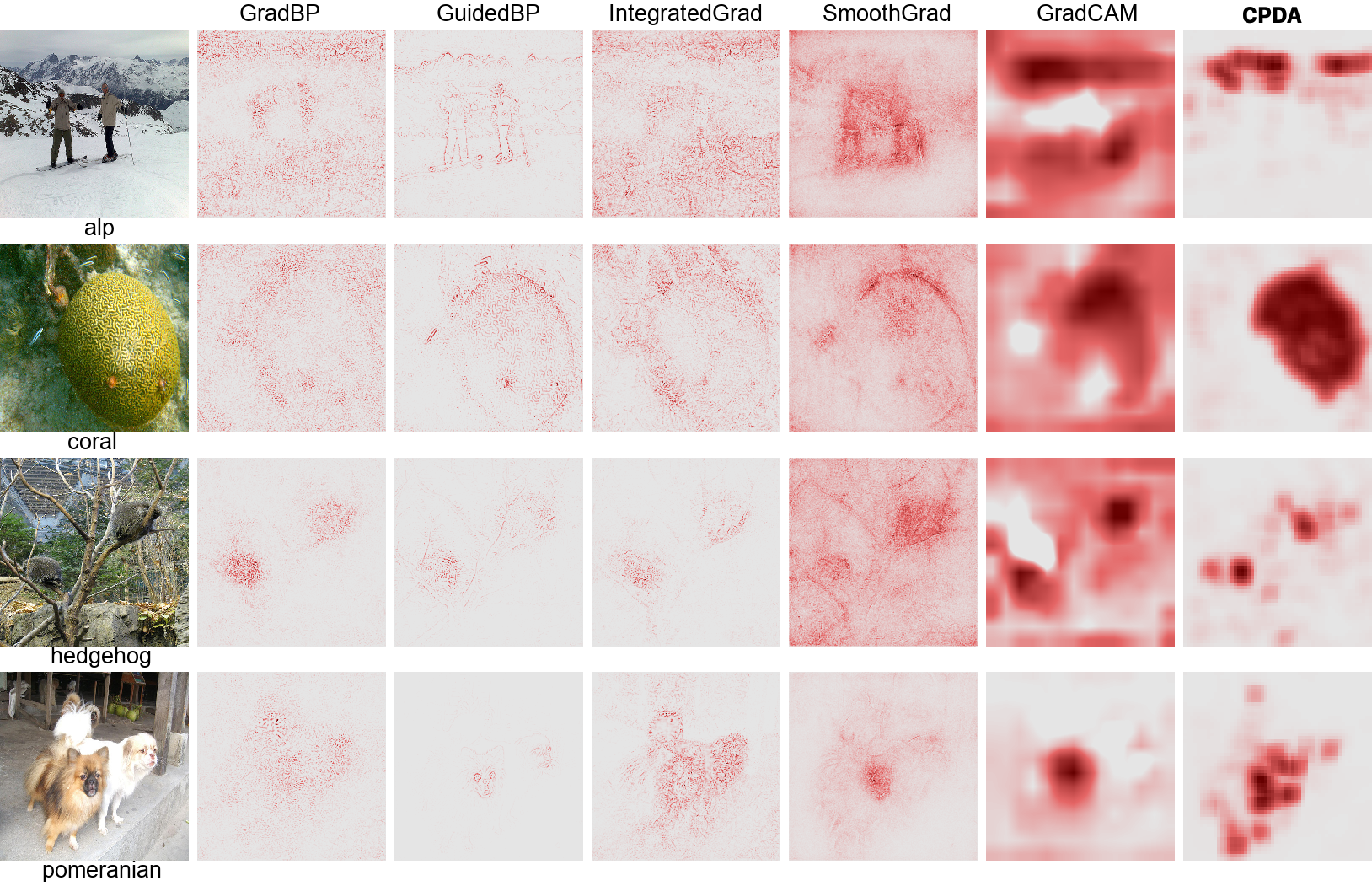}}
\caption{Explanations created by model-aware methods.} \label{fig:expla_maware}
\end{figure}

We also compare our method with popular model-aware saliency methods, namely, Gradients \cite{Simonyan2013DeepIC}, Guided Backpropagation \cite{springenberg2014striving}, IntergradtedGrad \cite{sundararajan2017axiomatic}, SmoothGrad \cite{smilkov2017smoothgrad}, Grad CAM  \cite{selvaraju2017grad}. In Figure \ref{fig:expla_maware}, for each image, we create saliency maps for the class that is given under the image in the figure. CPDA in the first row can identify pixels on the alp class. Although our method is model agnostic, the created explanations for some instances are competitive compared with the ones by model-aware methods.

\begin{figure}[h]
\centerline{\includegraphics[scale=0.6]{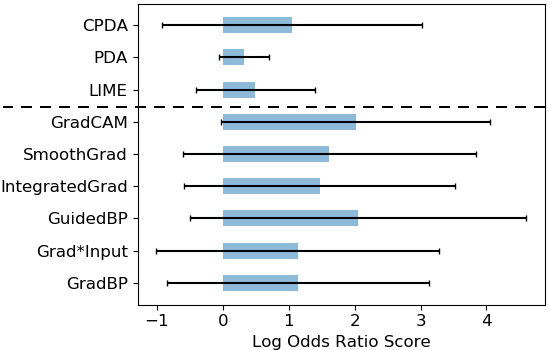}}
\caption{Quantitative Evaluation of Saliency Maps: the bars are means and the lines mean standard deviations.} \label{fig:quant_eval}
\end{figure}

We also quantitatively evaluate the created saliency maps. Given an image classification on VGG16 with an output of the $i$-th class $p^i$, we create a saliency map for this class $S^i$, identify the position with the maximum of $S^i$ and perturb the original image at the identified position with a 9$\times$9 patch of a constant mean value. In this way, the pixel most relevant to the $i$-th class is removed. The corresponding output becomes $q^i$. The Log Odds Ratio is defined as $log(\frac{p^i/(1-p^i)}{q^i/(1-q^i)})$. If the $S^i$ accurately identifies the relevant pixels, the corresponding output after the perturbation becomes smaller and Log Odds Ratio is high. The scores will be averaged on 1k validation images. Since the ground-truth explanations are unknown, the Log Odds Ratio score is applied to evaluate saliency maps approximately. The evaluation results are shown in Figure \ref{fig:quant_eval}. CPDA outperforms LIME and PDA. Compared with model-aware methods, CPDA also shows a comparable score to the approach apparoch (GradBP).

\begin{figure}[t] 
  \centering 
  \subfigure[Visualizing the effect of the patch size]{ 
    \label{fig:subfig:size} 
    \includegraphics[scale=0.17]{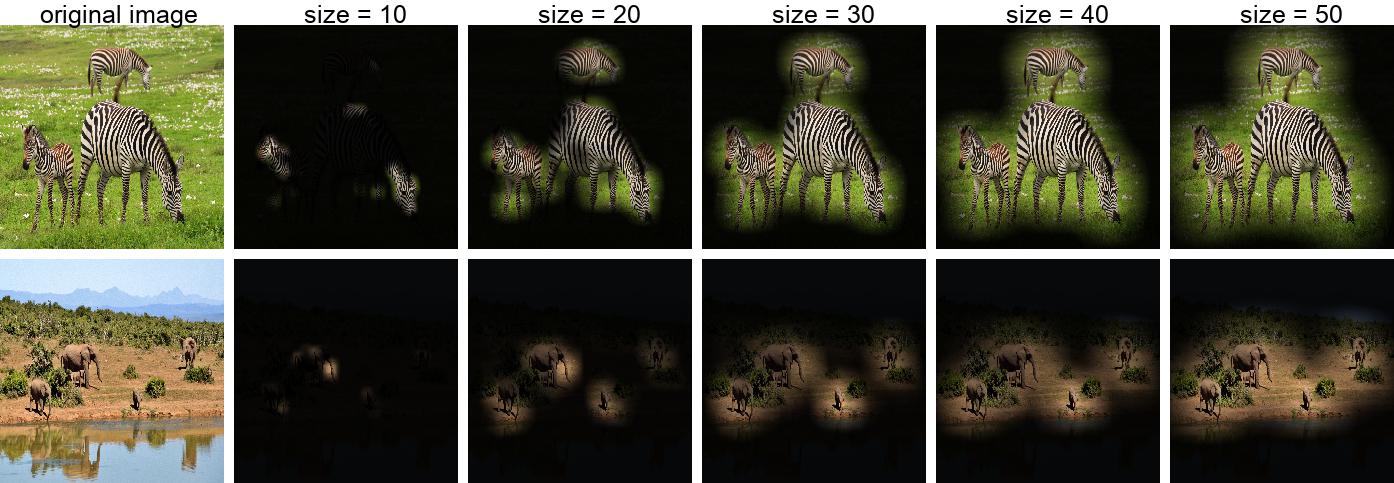}} 
  \subfigure[Visualizing the effect of the stride]{ 
    \label{fig:subfig:stride} 
    \includegraphics[scale=0.17]{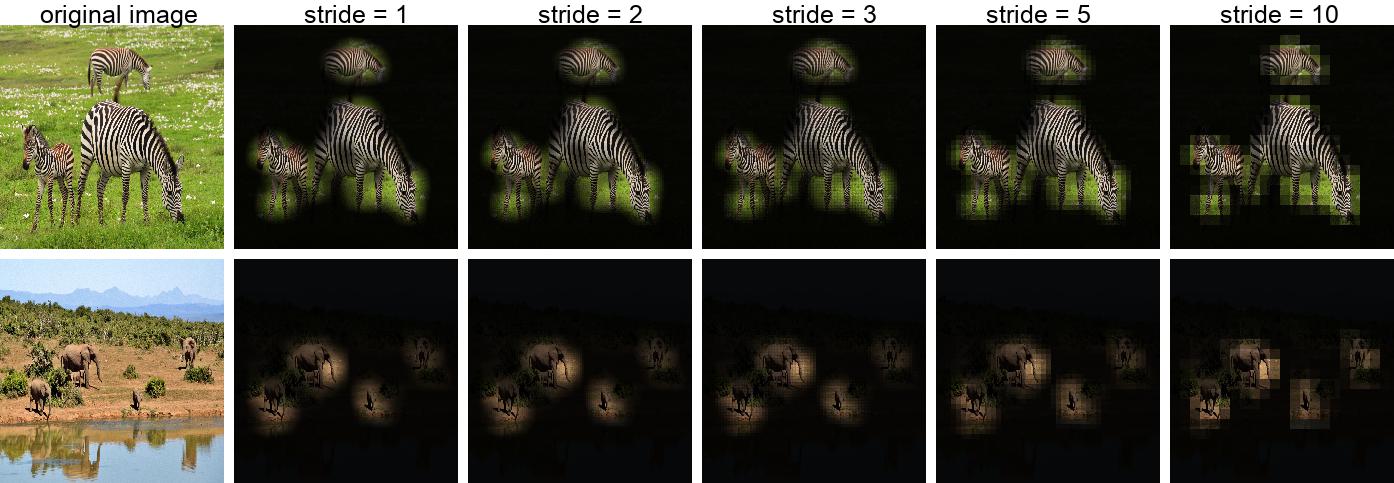}} 
  \caption{This figure shows the sensitivity of hyper-parameters in CPDA. The stride has almost no effect on the final results. The smooth effect of saliency maps depends on the clipped patch size.} \label{fig:hyper} 
\end{figure}

\subsubsection{Effects of Hyper-parameters:}
Two hyper-parameters are involved in our method, namely, the size of image patches and the stride of clipping operations. Figure \ref{fig:subfig:size} shows the effect of the patch size has on the final saliency map. When the small patch size is applied, the resulted saliency maps highlight the most discriminative visual features, e.g., the strip texture on zebra's heads and the long nose of elephants. When increasing the patch size, we can get a smoother map until the saliency map gets very blurry. The optimal patch size depends on the size of the target objects in the image. The size of 20 works well empirically in practice. Regardless of the patch size, the highlighted visual features always make sense to support the classification decision.

The resolution of the saliency map depends on the stride size. With an overlapping stride of 1, we can obtain pixel-wise accurate saliency maps. However, a small stride requires more forwarding inferences to compute the saliency map. The choice of stride is a trade-off between the resolution of saliency maps and computational cost. Figure \ref{fig:subfig:stride} shows that our approach is not sensitive to the stride. Too large stride may lead to a visual ``sawtooth'' effect, which makes the explanation less trustable. With a stride varying from 1 to 5, the visual ``sawtooth'' effect can be ignored. The stride of 5 is chosen in our experiments.

\begin{figure*}[t]
\centerline{\includegraphics[scale=0.18]{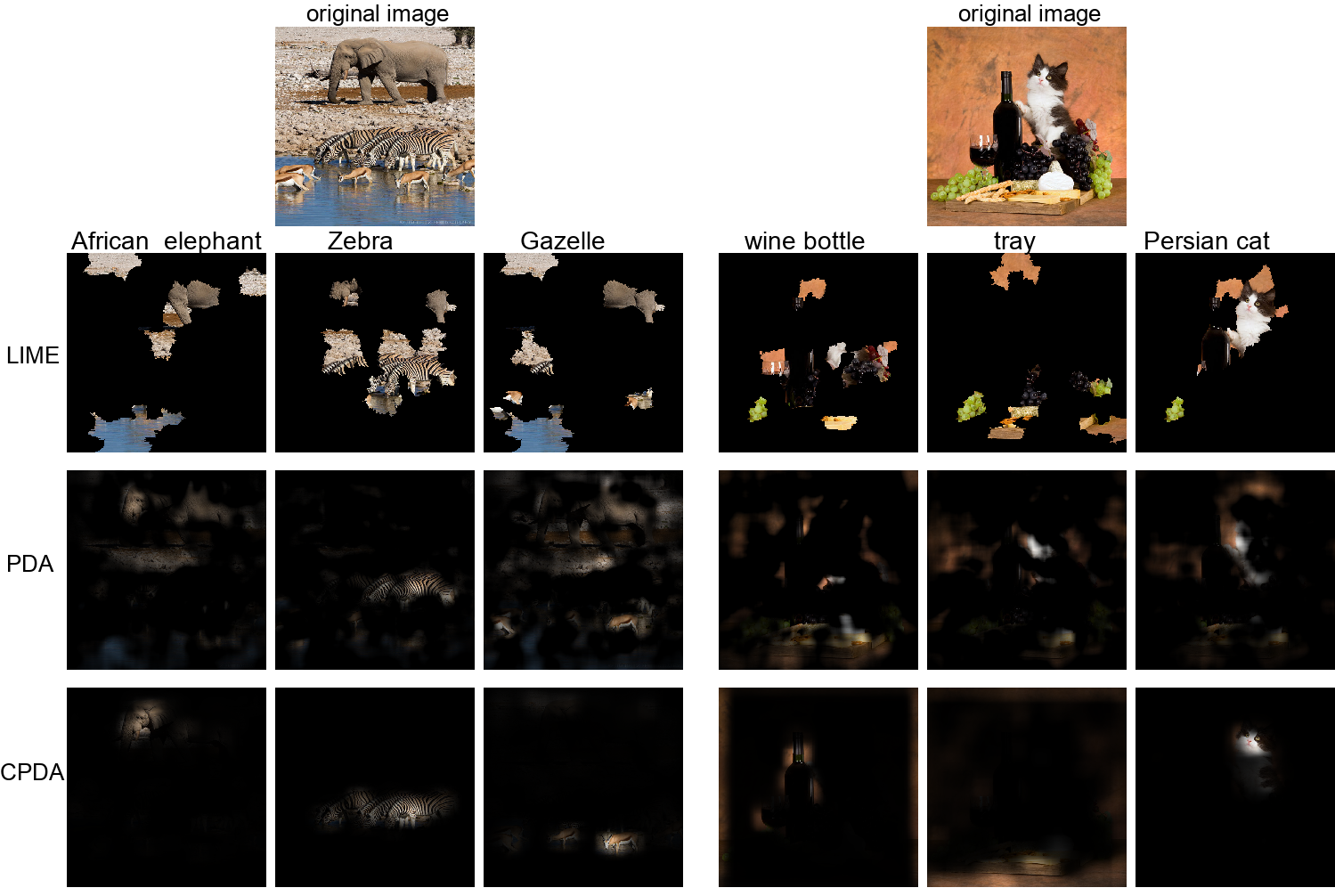}}
\caption{The figure shows the class-discriminativity of explanations. For each predicted class, the different methods are applied to produce an explanation visualizing relevant visual features. Please zoom in on the electronic version.} \label{fig:cls_dis}
\end{figure*}

\subsubsection{Class-discriminativity and Negative Evidence of Saliency maps:}
In the images with objects from a single class, the salient objects are often the objects relevant to the target class. Although saliency maps show visually recognizable target objects, saliency methods may only emphasize salient objects in images instead of identifying the relevant objects. Hence, a desired property of explanations is class-discriminativity. The image with objects from different classes is classified, where the classification is not saturated. The output probabilities are distributed across different classes. For each class, the saliency map is supposed to only highlight the class-relevant visual features instead of all salient ones.

Figure \ref{fig:cls_dis} shows the class-discriminativity of LIME, PDA and our CPDA. LIME trains a classifier to locally approximate the output of the model, which is distributed. With the parameters of the surrogate model, LIME can identify the super-pixels relevant to different classes. In the case of images with multi-class objects, the explanations of some classes often include irrelevant features. As shown in Figure \ref{fig:cls_dis}, the class of \textit{cat} is well explained with pixels of the cat object, and the explanations for other classes are not meaningful, e.g., \textit{wine bottle}. The explanations produced by PDA are not always understandable for all relevant classes. For instance, the identified visual features from the image in the second row for the classes (\textit{wine bottle, tray, Persian cat}) are not relevant, respectively. Our method CPDA is class-discriminative. It accurately identifies the class-relevant visual features to explain the output score of each class.

The quality of negative evidence of saliency maps is another indicator to evaluate saliency methods. The local surrogate classifier trained by LIME is explainable, which can identify which super-pixels make negative contributions. When marginalizing a small image patch in PDA, the increased classification score indicates that the marginalized image patch is the negative evidence to the classification. Since the classification can be saturated, the marginalization of relevant features not necessarily leads to an increase of output score. Therefore, the PDA does not work well to identify negative evidence. In our method CPDA, the contextual visual features are identified as negative evidence when $R_{\backslash i} = f(\pmb{x}) - f(x_i) < 0$. The contextual information is often treated together. Hence, irrelevant features receive similar negative relevance values.

\begin{figure}[t]
\centerline{\includegraphics[scale=0.17]{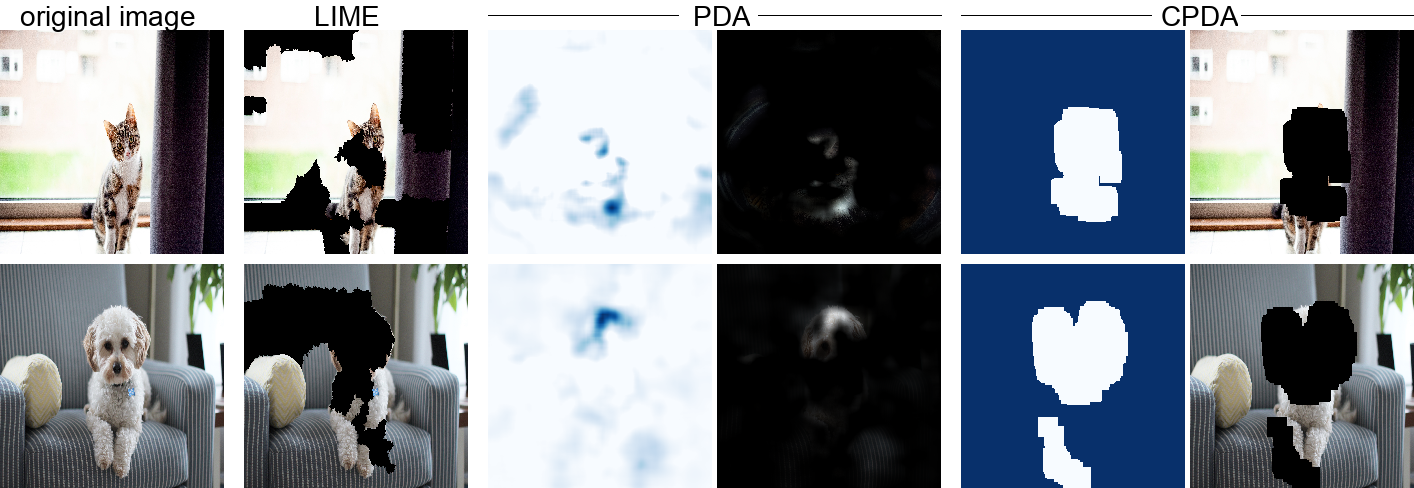}}
\caption{Negative evidences in classification explanations.} \label{fig:nega_exp}
\end{figure}

Figure \ref{fig:nega_exp} shows negative evidence of classifications identified by different methods. The visual features with strong negative evidence are disclosed, and the non-negative evidence is blocked out. For PDA and CPDA, saliency maps of negative evidence are also visualized, where blue means negative relevance value. The figure shows LIME and CPDA identify the contextual visual features as the reason why the classification confidence is not higher. The features with negative evidence in PDA are part of the target object instead of contextual distracting visual features. In the process of computing the marginalization of image patches, the classifier is applied to classify many constructed samples, which are not from the training data distribution. Hence, the computed effect is not unreliable because of the vulnerability of neural networks \cite{szegedy2013intriguing}.

\subsubsection{Efficiency and Consistency of Methods:}
The efficiency of model-agnostic saliency methods is non-ignorable. Since they have no access to the parameters of classifiers, the algorithms have to make many times inferences to obtain enough information for identifying the relevance of input features. Additionally, the consistency of produced explanations is another important indicator to evaluate saliency methods. The consistent explanations make more sense to understand the classifier and gain more trust from end-users.

For a single instance, LIME trains a local surrogate classifier to approximate the local classification boundary of the target classifier. Although the surrogate classifier takes the superpixels as individual features, LIME takes several minutes to produce an explanation for a classification. The explanations produced by LIME are inconsistent. The sources of the inconsistency could be the construction of instance variants, the type of surrogate classifiers, and the optimization process of the surrogate classifiers.

PDA considers patches of an image as individual features to produce explanations. For each image patch, the method PDA first models the conditional distribution $p(x_i | \mathring{U}(x_i))$ and samples the patches from the distribution. The modeling and sampling processes are computationally expensive. Given an input image of size $n \times n$ and a patch of size $k \times k$, $S(n-k+1)^2$ times forward passes are required to evaluate the relevance of all pixels, where $S$ means the number of samples from the distribution $p(x_i | \mathring{U}(x_i))$. Tens of minutes are required to produce an explanation.

\begin{table}[t]
\begin{center}
{\caption{The table lists the averaged time of producing one explanation for a single classification on different models.} \label{tab:effi}}
\begin{tabular}{cccc}
\toprule
Model &  LIME & PDA & CPDA \\
\midrule
AlexNet \tiny{CPU} & 5.15 \textit{min} & 380.82 \textit{min} & 1.81 \textit{min} \\
\midrule
AlexNet &  4.03 \textit{min} & 20.71 \textit{min} & 10.82 \textit{sec} \\
VGG16 &  5.17 \textit{min} & 86.62 \textit{min} & 42.68 \textit{sec} \\
Inception\_V3 &  6.06 \textit{min} & 184.67  \textit{min} & 93.34 \textit{sec} \\
ResNet &  4.46  \textit{min} & 97.21 \textit{min} & 49.10 \textit{sec} \\
DenseNet &  3.36 \textit{min} & 61.25 \textit{min} & 30.39 \textit{sec} \\
\bottomrule
\end{tabular}
\end{center}
\end{table}

Compared to LIME, CPDA requires neither the construction of noisy variant instances nor an extra classifier. Differently from PDA, CPDA does not have to model the conditional distribution of image patches. We classify image patches directly to identify the relevance of contextual pixels. With the same setting as in PDA, only $(\frac{(n-k+1)}{s})^2$ times inferences are required to produce an explanation using CPDA, where $k$ is the size of image patches, and $s$ is the stride. CPDA is $S*s^2$ times faster than PDA.  It only takes seconds or minutes to produce an explanation when s = 5. Table \ref{tab:effi} lists the averaged time to produce a single explanation. The first row shows the time of running the methods on AlexNet on a CPU (3,1 GHz Intel Core), while other rows show the time on a single GPU (Tesla K80).

\section{Related Work}
\label{sec:related}
The vanilla Gradient \cite{Simonyan2013DeepIC} identifies the gradients of an output with respect to inputs as saliency values. Guided Backpropagation \cite{springenberg2014striving} combines the Deconv visualization and vanilla Gradient by further discarding negative gradients during backpropagation. Integrated Gradients \cite{sundararajan2017axiomatic} computes the average gradient when the input varies linearly from a reference point to the original input. SmoothGrad \cite{smilkov2017smoothgrad} computes the average gradient under different perturbations to produce a sharper saliency map. Another method Grad-CAM produces class-discriminative saliency maps, which is efficient and widely applied \cite{selvaraju2017grad}. 

Differently from saliency methods above, model-agnostic saliency methods work without accessing to the model parameters \cite{guidotti2018survey}. For instance, LIME trains a small classifier with interpretable weights (e.g., linear classifier) to approximate the local decision boundary of a deep classifier \cite{ribeiro2016should}. For each classification decision, this approach requires to train a new classifier, which is inefficient. Given a single classification, the optimization of the surrogate classifier could end with different parameters, which leads to inconsistent explanations. When explaining a classification using different classifiers, LIME also produces different explanations. The work \cite{lundberg2017unified} presents a unified framework for interpreting predictions, SHAP (Shapley Additive exPlanations). They propose a model-agnostic approximation of SHAP value, called Kernel SHAP, which requires prohibitive computational cost when applied to real-world high-resolution images. The work \cite{fong2017interpretable} generates meaningful perturbation as explanations, which requires no information about model architectures. However, the generation process requires gradients of the model output with respect to the input.

PDA is another model-agnostic method to analyze classifiers. As a probabilistic sound methodology, PDA has received increased attention. PDA identifies the relevance of input features by marginalizing the effect of this feature and computing their prediction difference \cite{robnik2008explaining}. PDA faces many challenges when it is applied to high-dimensional data. It requires many forward inferences, which is computationally expensive \cite{zintgraf2017visualizing}. To reduce the times of inferences, the work of \cite{wei2018explain} analyzes the image in the super-pixel level. In this work, we present that PDA also suffers from model saturation and proposes a new approach to overcome the limitation.

\section{Conclusion}
In this work, we propose the Contextual Prediction Difference Analysis, an efficient, consistent, and model-agnostic saliency method. The empirical experiments illustrate the superiority of the proposed method. However, CPDA requires that classifiers are able to classify an instance on the part of features. The limitation will be further investigated in future work.

\bibliographystyle{splncs04}
\bibliography{samplepaper}

\end{document}